\pdfoutput=1

\documentclass[11pt]{article}

\usepackage{EMNLP2022}

\usepackage{times}
\usepackage{latexsym}

\usepackage[T1]{fontenc}

\usepackage[utf8]{inputenc}

\usepackage{microtype}

\usepackage{inconsolata}

\usepackage{booktabs}
\usepackage{multirow}
\usepackage{graphicx}
\usepackage{subcaption}
\usepackage{amsmath}
\usepackage{array}
\newcolumntype{T}{>{\tiny}l} %
\newcolumntype{P}[1]{>{\centering\arraybackslash}p{#1}} %

\usepackage[normalem]{ulem}
\usepackage{contour}

\contourlength{0.8pt}
\newcommand{\ul}[1]{%
  \uline{\phantom{#1}}%
  \llap{\contour{white}{#1}}%
}

\usepackage{placeins}
\usepackage{afterpage}

\usepackage{amsmath}
\DeclareMathOperator*{\argmax}{arg\,max}

\newcommand\ti[1]{\textit{#1}}

\newcommand\tf[1]{\textbf{#1}}

\title{Don't Prompt, Search! \\Mining-based Zero-Shot Learning with Language Models}

\newcommand{\affilsup}[1]{\rlap{\textsuperscript{\normalfont#1}}}
\author{
    Mozes van de Kar\affilsup{1} \qquad
    Mengzhou Xia\affilsup{2} \qquad
    Danqi Chen\affilsup{2} \qquad
    Mikel Artetxe\affilsup{3} \\
    $^1$University of Amsterdam \qquad
    $^2$Princeton University \qquad
    $^3$Meta AI \\
    \texttt{mozesvandekar@gmail.com} \qquad
    \texttt{\{mengzhou,danqic\}@cs.princeton.edu} \\ %
    \texttt{artetxe@meta.com}
}

\begin{document}
\maketitle
\begin{abstract}

Masked language models like BERT can perform text classification in a zero-shot fashion by reformulating downstream tasks as text infilling. However, this approach is highly sensitive to the template used to prompt the model, yet practitioners are blind when designing them in strict zero-shot settings. In this paper, we propose an alternative mining-based approach for zero-shot learning. Instead of prompting language models, we use regular expressions to mine labeled examples\footnote{We use `labeled examples' throughout the paper to denote the examples that match regex-based patterns of different labels. They are \ti{weakly-supervised} and can be noisy. } from unlabeled corpora, which can optionally be filtered through prompting, and used to finetune a pretrained model. Our method is more flexible and interpretable than prompting, and outperforms it on a wide range of tasks when using comparable templates. Our results suggest that the success of prompting can partly be explained by the model being exposed to similar examples during pretraining, which can be directly retrieved through regular expressions.
\end{abstract}

\section{Introduction}
\label{sec:introduction}

Recent work has obtained strong zero-shot results by prompting language models \citep{language2020brown,chowdhery2022palm}.
As formalized by \citet{schick-schutze-2021-exploiting}, the core idea is to reformulate text classification as language modeling using a \textit{pattern} and a \textit{verbalizer}. Given the input space $X$, the output space $C$ and the space of possible strings $V^*$, the pattern $t: X \rightarrow V^*$ maps each input into a string with a masked span, whereas the verbalizer $v: C \rightarrow V^*$ maps each label into a string. A language model can then be used for zero-shot classification by picking the most likely completion for the masked text $\argmax_{c \in C} p(v(c) \mid t(x))$.\footnote{We focus on masked language models, and allow multi-token verbalizers through autoregressive decoding (see \S\ref{sec:settings}). Left-to-right language models also fit the framework by placing the mask at the end or scoring the full populated prompt. %
}
In few-shot settings, better results can be obtained by prepending a few labeled examples \cite{language2020brown}, or using them in some form of fine-tuning \cite{schick-schutze-2021-exploiting,gao-etal-2021-making}.

\begin{figure}[t]
\includegraphics[width=0.90\linewidth]{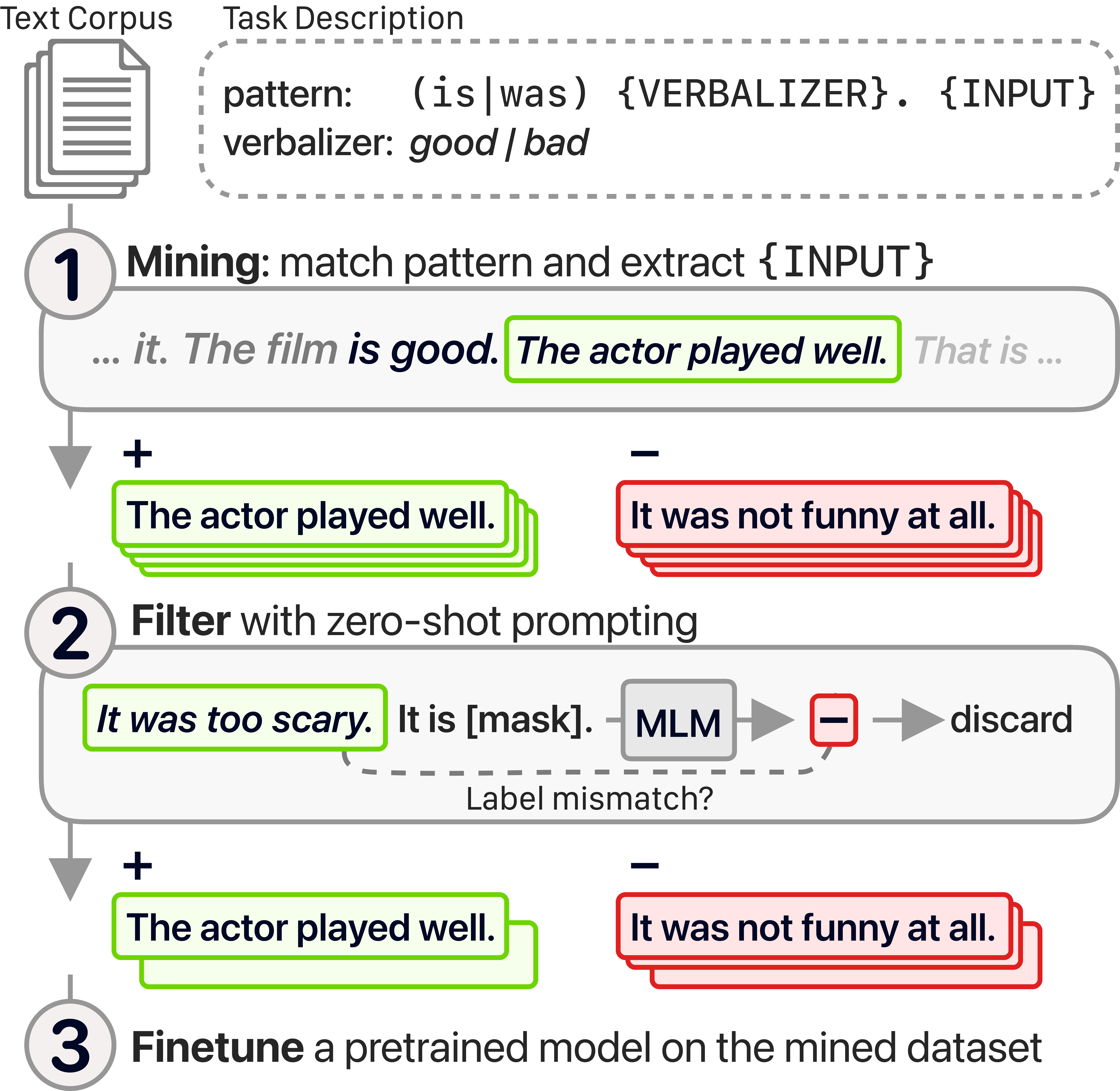}
\centering
\caption{
\textbf{Proposed method.}
1) We mine labeled examples from a text corpus with regex-based patterns.
2) Optionally, we filter examples for which zero-shot prompting predicts a different label.
3) We finetune a pretrained language model with a classification head.
}
\label{fig:method_overview}
\end{figure}

\begin{table*}[t]
\begin{center}
\begin{small}
\resizebox{\textwidth}{!}{
\begin{tabular}{lll}
\toprule
\textbf{Task} & \textbf{Prompting pattern} & \textbf{Mining pattern} \\
\midrule
Sentiment & \texttt{\{INPUT\}. It was \{VERBALIZER\}}. & \texttt{(is|was) \{VERBALIZER\}*. \{INPUT\}} \\
Topic class. & \texttt{\{INPUT\}. It is about \{VERBALIZER\}}. & \texttt{\{VERBALIZER\}*. \{INPUT\}} \\
NLI & \texttt{\{INPUT:HYP\} \{VERBALIZER\}, \{INPUT:PREM\}} & \texttt{\{INPUT:HYP\} \{VERBALIZER\}, \{INPUT:PREM\}} \\
\bottomrule
\end{tabular}
}
\end{small}
\end{center}
\caption{\textbf{Patterns.} \texttt{\{VERBALIZER\}} is replaced with the verbalizers in Table \ref{tab:verbalizers-short}. For mining, \texttt{*.} captures everything up to a sentence boundary, and \texttt{\{INPUT\}}, \texttt{\{INPUT:HYP\}} and \texttt{\{INPUT:PREM\}} capture a single sentence.
}
\label{tab:patterns}
\end{table*}

\begin{table}[t]
\begin{center}
\begin{small}
\addtolength{\tabcolsep}{-1.0pt}
\begin{tabular}{lll}
\toprule
\textbf{Task} & \textbf{Lbl} & \textbf{Verbalizers} \\
\midrule
\multirow{2}{*}{Sent.}
& Pos. & \ul{good}, great, awesome, incredible \\
& Neg. & \ul{bad}, awful, terrible, horrible \\
\midrule
\multirow{6}{*}{NLI}
& Ent. & \ul{Yes}, Therefore, Thus, Accordingly, \\
&& Hence, \textit{For this reason} \\
& Con. & \ul{No}, However, But, \textit{On the contrary}, \\
&& \textit{In contrast} \\
& Neu. & \ul{Maybe}, Also, Furthermore, Secondly, \\
&& Additionally, Moreover, \textit{In addition} \\
\bottomrule
\end{tabular}%
\end{small}
\end{center}
\caption{
\textbf{Verbalizers for sentiment classification and NLI.} See Table~\ref{tab:verbalizers-full} for verbalizers used in topic classification. When using a single verbalizer, we choose the one underlined.
Multi-token verbalizers are in italic.
Lbl: label, Ent./Con./Neu: entailment, contradiction, neutral. 
}
\label{tab:verbalizers-short}
\end{table}

\looseness=-1
However, prompting is known to be sensitive to the choice of the pattern and the verbalizer, yet practitioners are blind when designing them in true zero-shot settings \cite{can2020jiang,perez2021true}.
Connected to that, subtle phenomena like the surface form competition \cite{holtzman-etal-2021-surface} have a large impact on performance.
Recent work has tried to mitigate these issues through calibration \cite{calibrate2021zhao}, prompt combination  \cite{schick-schutze-2021-exploiting,lester-etal-2021-power,zhou2022prompt} or automatic prompt generation \cite{shin2020autoprompt,gao-etal-2021-making}.
At the same time, there is still not a principled understanding of how language models become few-shot learners, with recent work analyzing the role of the pretraining data \cite{chan2022data} or the input-output mapping of in-context demonstrations \cite{min2022rethinking}.

In this paper, we propose an alternative approach to zero-shot learning that is more flexible and interpretable than prompting, while obtaining stronger results in our experiments. Similar to prompting, our method requires a pretrained language model, pattern, and verbalizer, in addition to an unlabeled corpus (e.g., the one used for pretraining). As illustrated in Figure \ref{fig:method_overview}, our approach works by using the pattern and verbalizer to mine labeled examples from the corpus through regular expressions, and leveraging them as supervision to finetune the pretrained language model. This allows to naturally combine multiple patterns and verbalizers for each task, while providing a signal to interactively design them by looking at the mined examples. In addition, we show that better results are obtained by filtering the mined examples through prompting.

\looseness=-1
Experiments in sentiment analysis, topic classification and natural language inference (NLI) confirm the effectiveness of our approach, which outperforms prompting by a large margin when using the exact same verbalizers and comparable patterns. Our results offer a new perspective on how language models can perform downstream tasks in a zero-shot fashion, showing that similar examples often exist in the pretraining corpus, which can be directly retrieved through simple extraction patterns.

\section{Proposed Method}
\label{sec:method}

As shown in Figure \ref{fig:method_overview}, our method has three steps:
\vspace{-0.5em}
\paragraph{Mine.}
We first use the pattern and a set of verbalizers to extract labeled examples from the corpus. To that end, we define patterns that are filled with verbalizers and expanded into regular expressions. For instance, the pattern and verbalizer in Figure \ref{fig:method_overview} would extract every sentence following \textit{``is good.''} or \textit{``was good.''} as an example of the positive class, and every sentence following \textit{``is bad.''} or \textit{``was bad.''} as an example of the negative class. In practice, the patterns that we define are comparable to the ones used for prompting, and the verbalizers are exactly the same (see Tables \ref{tab:patterns} and \ref{tab:verbalizers-short}). Appendix \ref{sec:appendix-detailed-method} gives more details on how we expand patterns into regular expressions. While prior work in prompting typically uses a single verbalizer per class, our approach allows to naturally combine examples mined through multiple verbalizers in a single dataset. So as to mitigate class imbalance and keep the mined dataset to a reasonable size, we mine a maximum of 40k examples per class after balancing across the different verbalizers. 

\vspace{-0.5em}
\paragraph{Filter.}
As an optional second step, we explore automatically removing noisy examples from the mined data. To that end, we classify the mined examples using zero-shot prompting, and remove examples for which the predicted and the mined label do not match. This filtering step is reliant on the performance of prompting, and we only remove 10\% of the mismatching examples for which zero-shot prompting is the most confident. 

\vspace{-0.5em}
\paragraph{Finetune.}
Finally, we use the mined dataset to finetune a pretrained language model in the standard supervised fashion~\cite{devlin-etal-2019-bert}, learning a new classification head.

\begin{table*}[t]
\begin{center}
\begin{small}
\addtolength{\tabcolsep}{-2.5pt}
\begin{tabular}{cclccccccccccccccccc}
\toprule
&&&& \multicolumn{5}{c}{\textbf{Sentiment analysis}} && \multicolumn{3}{c}{\textbf{Topic class.}} && \multicolumn{4}{c}{\textbf{NLI}} && \multirow{2}{*}{\textbf{avg}}      \\
\cmidrule{5-9} \cmidrule{11-13} \cmidrule{15-18}
                    &&&& amz        & imd       & mr      & sst    & ylp    && agn           & dbp            & yah           && mnl & qnl & rte  & snl &&  \\
\midrule
Full-shot  && Fine-tuning    && 97.1        & 95.7       & 88.8    & 94.4    & 95.0    && 95.1           & 99.3           & 76.8            && 78.5                   & 92.6 & 67.6 & 90.5 && 89.3         \\
\midrule
\multirow{3}{*}{Zero-shot}

&& Prompting && 81.5 & 78.4 & 71.1 & 77.4 & 81.9 && 34.0 & 36.4 & 28.2 && 47.1 & 50.8 & 52.3 & 39.6 && 56.6 \\
&& \textit{\quad w/ multi verb.} && 83.5 & 81.8 & 78.3 & 81.9 & 83.1 && 54.6 & 51.1 & 34.1 && 46.5 & \bf 58.2 & 61.4 & 44.1 && 63.2 \\

&& Proposed method             && \bf 92.0        & \bf 86.7       & \bf 80.5    & \bf 85.6    & \bf 92.0    && \bf 79.2           & \bf 80.4           & \bf 56.1            && \bf 50.4                    & 53.2 & \bf 62.6 & \bf 46.0 && \bf 72.0        \\
\bottomrule
\end{tabular}
\end{small}
\end{center}
\caption{
\textbf{Main results (accuracy).}
All systems are based on RoBERTa-base, and all zero-shot systems use comparable patterns (see Table \ref{tab:patterns}). We report average accuracy across 3 runs for all systems except prompting. w/ multi verb.: prompting with different sets of verbalizers (Table~\ref{tab:verbalizers-full}) and averaging the probabilities.   
}
\label{tab:results}
\end{table*}

\section{Experimental Settings}
\label{sec:settings}

\paragraph{Tasks.}
We evaluate on three types of tasks: \textit{binary sentiment analysis} on Amazon \cite{dataset_amazon_yahoo_dbpedia_yelp_agnews}, IMDb \cite{dataset_imdb}, MR \cite{dataset_mr}, SST-2 \cite{dataset_sst2} and Yelp \cite{dataset_amazon_yahoo_dbpedia_yelp_agnews},
\textit{topic classification} on AG News \cite{dataset_amazon_yahoo_dbpedia_yelp_agnews}, DBPedia \cite{dataset_amazon_yahoo_dbpedia_yelp_agnews} and Yahoo Topics\footnote{The Yahoo Answers dataset was downloaded by, and access was limited to, the University of Amsterdam, where all experiments were carried out.} \cite{dataset_amazon_yahoo_dbpedia_yelp_agnews}, and \textit{NLI} on MNLI \cite{williams-etal-2018-broad}, QNLI \cite{dataset_qnli}, RTE \cite{dataset_rte1, dataset_rte2,dataset_rte3,dataset_rte4} and SNLI \cite{dataset_snli}.
We report accuracy on the test set when available, falling back to the validation set for SST-2, MNLI, RTE and QNLI.
For all systems involving fine-tuning, we report the average across 3 runs with different random seeds.
We ran all development experiments on SST-2 and AG News without any exhaustive hyperparameter exploration, and evaluate the rest of the tasks blindly.

\setlength{\leftmargini}{0.5cm}
\setlength{\leftmarginii}{0.5cm}

\paragraph{Approaches.} We compare the following methods in our experiments, using RoBERTa-base \cite{roberta} as the pretrained model in all cases:
\begin{itemize}
    \item \textbf{Full-shot fine-tuning}: We finetune RoBERTa on the original training set adding a new classification head. We train for 3 epochs with a batch size of 32. All the other hyperparameters follow \citet{roberta}. Refer to Appendix \ref{sec:appendix-detailed-setup} for more details.
    \item \textbf{Zero-shot prompting}: Standard prompting, described in \S\ref{sec:introduction}. Multi-token verbalizer probabilities are calculated autoregressively, picking the most likely token at each step \cite{just2020schick}.
    We report results using both a single verbalizer per class, as it is common in prior work, as well as multiple verbalizers per class, which is more comparable to our approach. For the latter, we combine the probabilities of each verbalizer by averaging.\footnote{We also tried summing or taking the maximum, which obtained similar results as shown in Appendix \ref{app:results}.}
    \item \textbf{Zero-shot mining}: Our proposed method, described in \S\ref{sec:method}. For the mining step, we use the first 100 shards from the C4 corpus \cite{2019t5c4}, which cover 9.8\% of the data. For the filtering step, we use single-verbalizer prompting to filter 10\% of the mislabeled examples. %
    For the fine-tuning step, we use the same settings as in the full-shot setup, except that we train for 5,000 steps with a dropout probability of 0.4.\footnote{
During development, we found that high dropout and early stopping help mitigating model overfitting caused by the misalignment between the mined and the true distribution.
However, evaluation on all tasks shows mixed results. We stick to the original setup with high dropout to be faithful to the rigorous zero-shot scenario, and report additional results with standard dropout in Appendix \ref{app:results}.
} 
To mitigate class imbalance, we form batches by first sampling the class for each instance from the uniform distribution, and then picking a random example from the mined data belonging to that class.
\end{itemize}

\begin{table}[t]
\begin{center}
\addtolength{\tabcolsep}{-1.5pt}
\resizebox{0.85\columnwidth}{!}{
\begin{tabular}{lcc}
\toprule
&  \textbf{Prompting} & \textbf{Mining} \\
\midrule
\textit{good} / \textit{bad} &  \textbf{78.1} & 72.0 \\
\textit{great} / \textit{awful} &  \textbf{82.3} & 82.1 \\
\textit{awesome} / \textit{terrible} & 82.3 & \textbf{83.9} \\
\textit{incredible} / \textit{horrible} & 83.1 & \textbf{87.3}  \\
\midrule
combined & 81.7 & \textbf{85.4} \\
\bottomrule
\end{tabular}
}
\end{center}
\caption{
\textbf{Average sentiment accuracy using different verbalizers.} We report mining results without filtering. More detailed results are provided in Table~\ref{tab:verbalizer-results-appendix}.
}
\label{tab:results-verbalizers}
\end{table}

\paragraph{Patterns and verbalizers.}
We use comparable patterns for prompting and mining  with the exact same verbalizers, which we report in Table \ref{tab:patterns} and \ref{tab:verbalizers-short}. These were designed without any experiment, simulating a zero-shot setting.
We design our patterns to capture sentences following a verbalizer, rather than sentences containing the verbalizer, as the resulting dataset would otherwise be trivial (solvable by detecting the presence of certain words).

\section{Results and Analysis}

\begin{table}[t]
\begin{center}
\begin{small}
\addtolength{\tabcolsep}{-2.5pt}
\begin{tabular}{@{}ll|l}
\toprule
\textbf{Pattern}                                      & \textbf{Verbalizer}  & \textbf{avg}  \\ \midrule
\textbf{Prompting}                           &                         &      \\
\verb|{VERBALIZER} stars: {INPUT}|           & \textit{5 / 1}          & 51.0 \\
\verb|{INPUT} I {VERBALIZER} it.|            & \textit{love / hate }   & 73.1 \\
\verb|{INPUT} It is {VERBALIZER}.|           & \textit{good / bad }    & \textbf{78.1} \\ \midrule
\textbf{Mining}                              &                         &      \\
\verb|{VERBALIZER} star*. {INPUT}|           &\textit{ 5 / 1}          & \textbf{72.1} \\
\verb|I {VERBALIZER}*. {INPUT}|              & \textit{love / hate}    & \textbf{85.5} \\
\texttt{(is|was) \{VERBALIZER\}*. \{INPUT\}} & \textit{good / bad}     & 72.0 \\ \bottomrule
\end{tabular}
\end{small}
\end{center}
\caption{\textbf{Average sentiment accuracy using different patterns and verbalizers.} We report mining results without filtering (more details are provided in Table~\ref{tab:template-sensitivity-full}).}
\label{tab:template-sensitivity}
\end{table}

\begin{table}[t]
\begin{center}
\resizebox{0.9\columnwidth}{!}{
\addtolength{\tabcolsep}{-2.5pt}
\begin{tabular}{lcccc}
\toprule
\textbf{Data} & \textbf{Filter} & \textbf{Sent.} & \textbf{Topic} & \textbf{NLI} \\
\midrule
Supervised & - & 94.2 & 90.4 & 82.3 \\ \midrule
\multirow{3}{*}{Mined}
& none & 85.4 & 71.5 & 52.0 \\
& prompting & 87.4 & 71.9 & 53.0 \\
& supervised$^\dagger$ & {91.4} & 85.5 & 63.8 \\
\bottomrule
\end{tabular}
}
\end{center}
\caption{
\textbf{Filtering results (average accuracy).}
$^\dagger$: uses mined data for training and another supervised classifier as the filter. This is not a zero-shot setting and serves as an upper limit for the results using a perfect filter. More detailed results are provided in Table~\ref{tab:complete-results-appendix}. 
}
\label{tab:results-filtering}
\end{table}

We next discuss our main findings and report additional results in Appendix \ref{app:results}.

\paragraph{Main results.} We report our main results in Table \ref{tab:results}. Our method outperforms prompting by 8.8 points on average, and the improvements are consistent across all tasks.

\paragraph{Effect of patterns and verbalizers.} Table \ref{tab:results-verbalizers} reports sentiment results using different verbalizers. Consistent with prior work, we find that both prompting and mining are highly sensitive to the choice of the verbalizer, yet combining them all roughly matches the results of the best performing one.
As shown in Table \ref{tab:template-sensitivity}, using different patterns has an even larger impact. Interestingly, patterns and verbalizers that do well with one approach do not necessarily do well with the other.

\paragraph{Effect of filtering.} Table \ref{tab:results-filtering} reports additional results using the full-shot systems for filtering, or not using any filtering at all.
We find that prompting-based filtering brings modest but consistent improvements across all types of tasks. 
We compare this to filtering out all examples with mismatching labels with the full-shot model, which results in much larger gains and approaches the performance of the fully supervised system for sentiment and topic classification tasks. This can be seen as an upper-limit of what could be reached with perfect filtering, which leaves ample room to improve our approach focusing on the filtering step alone.

\paragraph{Qualitative analysis.}
We manually assessed 20 mined examples for sentiment analysis and report some representative instances in Table \ref{tab:examples}. We find that the mined data covers many domains like finance and technology. Most examples are correct (\#1, \#3), but there are also instances with wrong labels (\#4). In addition, we find that 40\% of analyzed examples show weak or neutral sentiment (\#2). The impact of such irrelevant examples is unclear and worth of future study.

\begin{table}[t]
\begin{center}
\begin{small}
\addtolength{\tabcolsep}{-3pt}
\begin{tabular}{llp{60mm}}
\toprule
\# & \tf{Lbl} & \tf{Mined example} \\
\midrule
1 & Pos. & Do you have an idea of how broad your vocal range was? \\
2 & Pos. & Once home, we began priming. \\
3 & Neg. & People in Wall Street and other financial services firms should have paid more attention to the data. \\
4 & Neg. & So I bought this unit, which said it had the same technical features as the other brand, such as number of channels etc, and this one performed amazing!! \\
 \bottomrule
\end{tabular}
\end{small}
\end{center}
\caption{\textbf{Mined examples} for sentiment analysis. See more examples in Table~\ref{tab:sentiment-mined-examples-appendix} and mined NLI examples in Table~\ref{tab:nli-mined-examples-appendix}.
}
\label{tab:examples}
\end{table}

\section{Related work}

\looseness=-1
Recent work in zero-shot learning has explored a similar \textit{generate-filter-finetune} approach, but using large language models instead of mining to generate training data \cite{generating2021schick, wanli2022liu,meng2022generating,ye2022zerogen}. %
Mining-based approaches have a long tradition in information extraction \cite{automatically1996riloff, learning1999riloff}. However, to the best of our knowledge, we are the first to apply them for zero-shot learning as an alternative to prompting.
Instead of mining examples for the target task, \citet{Bansal2020SelfSupervisedMF} define task-agnostic pretraining objectives on unlabeled corpora.
Closer to our work, \citet{meng-etal-2020-text} mask label-indicative words in an unlabeled corpus, and train a model to predict their corresponding label.
Concurrent to our work, \citet{han2022orca} try locating a subset of the pretraining data that supports prompting in specific tasks. Finally, \citet{razeghi2022impact} show a strong correlation between performance on specific instances and the frequency of terms from those instances in the pretraining data.

\section{Conclusions}

In this work, we have shown that mining-based zero-shot learning outperforms prompting. Moreover, our approach shows headroom for further improvement by exploring filtering techniques. The flexibility of our approach enables additional directions like domain filtering, bootstrapping, and interactive pattern/verbalizer design, where practitioners would inspect a few mined examples and refine their patterns until they are satisfied.
In addition, our methods can serve as a partial explanation for why prompting works, showing that task-relevant examples are often present in the pretraining corpus in an explicit form, to the extent that they can be directly mined through simple regular expressions. Nevertheless, we believe that there can be other factors involved, as evidenced by the best patterns and verbalizers being different for mining and prompting, and we believe that delving deeper into the relation between pretraining data and prompting performance is an interesting future direction.

\section*{Limitations}

Developing zero-shot methods in a rigorous manner is challenging: the strict zero-shot scenario does not allow using annotated data except for the final evaluation, yet it is difficult to make development decisions without any signal. We decided to use AG News and SST-2 during development without any exhaustive hyperparameter exploration, and evaluate blindly in the rest of the tasks. At the same time, we designed all patterns and verbalizers without any experiment, based solely on our own intuition. We believe that the comparison between prompting and mining is fair as we used comparable patterns with the exact same verbalizers and pretrained model. However, it is possible that our patterns, verbalizers and/or hyperparameters are suboptimal, and better results could be obtained with either prompting or mining using other configurations.

An important limitation of our approach is that it can be difficult to design extraction patterns for certain tasks like multiple choice questions. However, prompting is known to suffer from a similar limitation, with certain tasks like WiC being difficult to formulate as language modeling and obtaining random chance performance \cite{language2020brown}.

Different from prompting, our approach requires an intermediate step after pretraining to mine data and finetune the model, which takes 2-7 hours using a single Nvidia Titan RTX GPU and 4 Intel Xeon CPUs. However, inference cost is similar or even faster than prompting, as our approach does not incur on any overhead for multi-token and multi-verbalizer setups.

\section*{Acknowledgements}
We thank Ves Stoyanov, Jingfei Du, Timo Schick and Sewon Min for their feedback. Mozes van de Kar received a travel grant from ELLIS and Qualcomm to attend the conference.

\bibliography{custom, litmaps, datasets,anthology}
\bibliographystyle{acl_natbib}

\appendix

\clearpage

\section{Pattern expansion for mining}
\label{sec:appendix-detailed-method}

For each class, examples are mined by filling in the pattern with the verbalizer and extracting sentences that match the filled-in pattern. The process of expanding the patterns into regular expressions is as follows. First, we replace \verb|{VERBALIZER}| with a capturing group containing all verbalizers separated by the alternation operator \texttt{|}.
For example, the verbalizer \textit{good, great, awesome} is expanded into \texttt{(good|great|awesome)}.
Finally, we replace the keywords described in Table \ref{tab:regex-language} with the corresponding regular expressions. The result is a regular expression containing capturing groups for extracting sentences in a case-insensitive fashion.

Note that we use a simplistic sentence definition in order to keep the regex manageable. Since we assume that a period always ends a sentence, this mistakenly interprets abbreviations as multiple sentences (e.g., \textit{``U.S.A.''} contains 3 sentences). To address this, we filter out mined sentences shorter than 4 characters.

\FloatBarrier

\begin{table}[t]
\begin{center}
\begin{small}
\begin{tabular}{lp{46mm}}
\toprule
\textbf{Keyword} & \textbf{Regex}                                                                                                                                                                       \\ \midrule
\verb|{VERBALIZER}|    & Replaced with the verbalizer                                                                                                                                                   \\ \midrule
\verb|*|         & regex: \verb|[^.!?]*?| \newline Greedily matches non-sentence-ending characters                                                                                                                            \\ \midrule
\verb|{INPUT}|    & regex: \verb|([^.!?]+[.!?]+)| \newline Matches a single sentence, extracted with the key ``INPUT''                       \\ \bottomrule
\end{tabular}
\end{small}
\end{center}
\caption{\textbf{Keywords that compile into regular expressions.} These keywords are used in the mining patterns and verbalizers. %
}
\label{tab:regex-language}
\end{table}

\section{Additional experimental details}
\label{sec:appendix-detailed-setup}

\paragraph{Patterns and verbalizers.} For each category of tasks we use the same mining pattern, as shown in Table \ref{tab:patterns}. 
The complete list of verbalizers for each task is given in Table \ref{tab:verbalizers-full}.
Tasks with the same classes share the same verbalizers. This means that all sentiment and NLI tasks have the same verbalizers.
Each topic classification task, however, has a unique set of verbalizers. 
Note that while SNLI and MNLI (3-way NLI) have the same verbalizers as RTE and QNLI (2-way NLI), the mined datasets do differ since 2-way NLI does not include a neutral class.

\paragraph{Hyperparameters.} Table \ref{tab:appendix-hyperparameters} shows the hyperparameters used for finetuning the RoBERTa-base model. All the other hyperparameters and classification head architecture follow \citet{roberta}.
We have two fine-tuning configurations, one for fine-tuning in the full-shot setting and one for zero-shot fine-tuning on the mined dataset. These configurations differ only in the maximum number of steps, dropout rate and batch sampler. 

\paragraph{Datasets.} We use Huggingface \cite{lhoest-etal-2021-datasets} for loading all evaluation datasets without any additional processing, except for MR which is detokenized using Moses scripts. We evaluate on the test set, falling back to the validation set for SST-2, MNLI, RTE and QNLI.

\section{Additional results}
\label{app:results}

Complete results for full-shot, prompting and mining are combined in Table \ref{tab:complete-results-appendix}. 
Results showing the effect of pattern and verbalizer choice on binary sentiment classification are presented in Table \ref{tab:verbalizer-results-appendix} and Table \ref{tab:template-sensitivity-full}, respectively.

As explained in the main text, development experiments were only conducted on AGNews and SST-2. On these tasks, we found that high regularization partially mitigates overfitting caused by the misalignment between the mined dataset and real dataset. However, this high regularization shows mixed results for non-development tasks. For full transparency, we compare these performance differences in Table \ref{tab:results-normal-dropout}, but, in the main text, we stick to the original setup with high dropout to be faithful to the rigorous zero-shot scenario.

For multi-verbalizer prompting, we combine the probabilities of each verbalizer with an aggregation function. Results for using the average, the max and the sum are shown in Table \ref{tab:results-prompting-agg}.

In Table \ref{tab:label-agreement} we show the agreement between the mined labels and the labels according to the filtering method, which in our experiments is either a full-shot finetuned model or single-verbalizer prompting.

Table \ref{tab:sentiment-mined-examples-appendix} and Table \ref{tab:nli-mined-examples-appendix} show a random sample of examples from the mined training dataset for respectively binary sentiment analysis and NLI. 
In the main text, Table \ref{tab:examples} shows a representative selection of examples for sentiment analysis. These examples were manually picked from the random sample in Table \ref{tab:sentiment-mined-examples-appendix}.

\begin{table*}[t]
\begin{center}
\begin{small}
\begin{tabular}{lll}
\toprule
\textbf{Task} & \textbf{Class} & \textbf{Verbalizers} \\
\midrule
\multirow{2}{*}{Sentiment}
& Positive & \ul{good}, great, {awesome, incredible} \\
& Negative & \ul{bad},{ awful, terrible, horrible} \\
\midrule
\multirow{4}{*}{AGNews}
& World & \ul{world}, foreign, global, Asia, Europe, China \\
& Sports & \ul{sports}, football, basketball, {tennis, soccer, baseball} \\
& Business & \ul{business}, stock, financial, profit, {economy, finance} \\
& Sci/Tech & \ul{technology}, science, research, chemical, iPhone, {smartphone} \\
\midrule
\multirow{14}{*}{DBPedia}
& Company & \ul{company}, business, manufacturer, \textit{operates in} \\
& Educational institution & \ul{school}, college, education, {university} \\
& Artist & \ul{artist}, writer, song, {composer} \\
& Athlete & \ul{sports}, runner, basketball, football \\
& Office holder & \ul{politics}, president, Senate, {politician} \\
& Mean of transportation & \ul{bus}, bike, car, train, ship, plane, {aircraft} \\
& Building & \ul{building}, office, house, {monument} \\
& Natural place & \ul{river}, forest hill, {nature} \\
& Village & \ul{town}, village, \textit{small population, small town} \\
& Animal & \ul{animal}, species, horse, dog, pet, {habitat} \\
& Plant & \ul{plant}, leaf, flower, {herb} \\
& Album & \ul{album},recording,  \textit{record company} \\
& Film & \ul{film}, movie, actor, {actress} \\
& Written work & \ul{written}, book,{ novel, poem} \\
\midrule
\multirow{14}{*}{Yahoo}
& Society \& Culture & \ul{culture}, holiday, {society} \\
& Science \& Mathematics & \ul{science}, technology, math, research \\
& Health & \ul{health}, body, exercise, \textit{stress relieve} \\
& Education \& Reference & \ul{school}, college, education, {university} \\
& Computers \& Internet & \ul{computer}, internet, keyboard, software \\
& Sports & \ul{sports}, football, basketball, game \\
& Business \& Finance  & \ul{business}, stock, financial, profit \\
& Entertainment \& Music  & \ul{film}, movie, actor, writer \\
& Family \& Relationships  & \ul{love}, family, father, mother \\
& Politics \& Government & \ul{politics}, president, Senate, {politician} \\
\midrule
\multirow{3}{*}{NLI}
& Entailment & \ul{Yes}, Therefore, Thus, Accordingly, Hence, \textit{For this reason} \\
& Contradiction & \ul{No}, However, But,\textit{ On the contrary, In contrast} \\
& Neutral & \ul{Maybe}, Also, Furthermore, Secondly, Additionally, Moreover, \textit{ In addition} \\
\bottomrule
\end{tabular}%
\end{small}
\end{center}
\caption{
\textbf{Verbalizers.}
When using a single verbalizer we choose the one underlined. In the multi-verbalizer setting we use all listed verbalizers. Sentiment includes Amazon, IMDB, MR, SST-2 and Yelp; NLI includes MNLI, QNLI, RTE and SNLI. Multi-token verbalizers are italic.
}
\label{tab:verbalizers-full}
\end{table*}

\begin{table*}[t]
\begin{center}
\begin{small}
\addtolength{\tabcolsep}{-2.5pt}
\begin{tabular}{@{}lP{30mm}P{30mm}@{}}
\toprule
\textbf{Parameter}                   & \textbf{full-shot}           & \textbf{zero-shot}                  \\ \midrule
Model              &  RoBERTa-base (123M) & RoBERTa-base (123M)                 \\
Model selection    &  last & last                         \\
Batch size         &  32 & 32                           \\
Optimizer          &  adam & adam                         \\
Learning rate      &  1.00e-05 & 1.00e-05                     \\
LR schedule        &  6\% warmup with linear decay & 6\% warmup with linear decay \\
Adam epsilon       &  1.00e-08 & 1.00e-08                     \\
Adam $\beta_1$       &  0.9 & 0.9                     \\
Adam $\beta_1$        &  0.999 & 0.999                     \\
Weight decay       &  0 & 0                            \\
Classifier dropout &  0 & 0                            \\
Attention dropout  &  0.1 & 0.1                          \\
Hidden dropout     & 0.1                 & 0.4                        \\
Max steps          & -                   & 5000                       \\
Max epochs         & 3                   & -                          \\
Batch sampler      & -                   & inverse class frequency\newline weighted sampling         \\ \bottomrule
\end{tabular}
\end{small}
\end{center}
\caption{\textbf{Hyperparameters} for full-shot finetuning and zero-shot finetuning with the mined dataset.}
\label{tab:appendix-hyperparameters}
\end{table*}

\begin{table*}[t]
\begin{center}
\begin{small}
\addtolength{\tabcolsep}{-2.5pt}
\begin{tabular}{@{}llll@{}Tlllll@{}Tl@{}Tl@{}Tl@{}T@{}}
\toprule
                              &  &  & \multicolumn{2}{c}{\textbf{Full-shot}}                    &  & \multicolumn{2}{c}{\textbf{Prompting}} &  & \multicolumn{8}{c}{\textbf{Mining}}                                                                                      \\ \cmidrule(lr){4-5} \cmidrule(lr){6-8} \cmidrule(l){10-17}
                              &  &  &  \multicolumn{2}{l}{fine-}  &  & single-            & multi-            &  & \multicolumn{2}{l}{single-} & \multicolumn{2}{l}{multi-} & \multicolumn{2}{l}{+ filter}  &               &               \\
                              & maj &  &   \multicolumn{2}{l}{tuning}  &  & verb               & verb              &  & \multicolumn{2}{l}{verb}    & \multicolumn{2}{l}{verb}   & \multicolumn{2}{l}{full-shot} & \multicolumn{2}{l@{}}{zero-shot} \\ \midrule
\textbf{Sentiment Analysis}   &  &  &                   &                     &  &                    &                   &  &              &              &              &             &                &              &               &               \\
Amazon                        & 50.0 &  & 97.1          & ±0.0                &  & 81.5               & 83.5              &  & 71.7         & ±2.2         & 90.6         & ±1.7        & 94.4           & ±0.2         & 92.0          & ±0.6          \\
IMDB                          & 50.0 &  & 95.7          & ±0.0                &  & 78.4               & 81.8              &  & 78.2         & ±0.5         & 83.0         & ±4.1        & 91.6           & ±0.4         & 86.7          & ±1.3          \\
MR                            & 50.0 &  & 88.8          & ±0.1                &  & 71.1               & 78.3              &  & 70.5         & ±1.0         & 77.7         & ±2.4        & 86.3           & ±0.7         & 80.5          & ±1.0          \\
SST-2                         & 50.9 &  & 94.4          & ±0.2                &  & 77.4               & 77.4              &  & 77.3         & ±1.4         & 83.8         & ±2.4        & 89.5           & ±0.7         & 85.6          & ±1.1          \\
Yelp                          & 50.0 &  & 95.0          & ±0.1                &  & 81.9               & 83.1              &  & 62.1         & ±2.3         & 92.0         & ±2.2        & 95.2           & ±0.6         & 92.0          & ±1.5          \\
\textbf{avg}                  & 50.2 &  & 94.2          &                     &  & 78.1               & 81.7              &  & 72.0         &              & 85.4         &             & 91.4           &              & 87.4          &               \\ \midrule
\textbf{Topic Classification} &  &  &                   &                     &  &                    &                   &  &              &              &              &             &                &              &               &               \\
AGNews                        & 25.0 &  & 95.1          & ±0.1                &  & 34.0               & 54.6              &  & 71.0         & ±0.7         & 78.4         & ±0.6        & 89.5           & ±0.2         & 79.2          & ±0.6          \\
DBPedia                       & 7.1 &  & 99.3           & ±0.0                &  & 36.4               & 51.1              &  & 63.7         & ±1.0         & 79.8         & ±0.0        & 97.1           & ±0.3         & 80.4          & ±0.4          \\
Yahoo                         & 10.0 &  & 76.8          & ±0.1                &  & 28.2               & 34.1              &  & 51.7         & ±0.5         & 56.3         & ±2.5        & 69.8           & ±0.1         & 56.1          & ±2.1          \\
\textbf{avg}                  & 14.0 &  & 90.4          &                     &  & 32.9               & 46.6              &  & 62.1         &              & 71.5         &             & 85.5           &              & 71.9          &               \\ \midrule
\textbf{NLI}                  &  &  &                   &                     &  &                    &                   &  &              &              &              &             &                &              &               &               \\
MNLI                          & 35.3 &  & 78.5          & 0.0                 &  & 47.1               & 46.5              &  & 48.2         & ±0.5         & 49.2         & ±0.5        & 65.5           & ±0.3         & 50.4          & ±0.4          \\
QNLI                          & 50.5 &  & 92.6          & ±0.1                &  & 50.8               & 58.2              &  & 51.6         & ±0.3         & 52.9         & ±1.2        & 70.2           & ±1.1         & 53.2          & ±0.6          \\
RTE                           & 52.7 &  & 67.6          & ±1.4                &  & 52.3               & 61.4              &  & 55.0         & ±0.2         & 61.1         & ±2.1        & 57.8           & ±1.3         & 62.6          & ±0.9          \\
SNLI                          & 34.3 &  & 90.5          & ±0.1                &  & 39.6               & 44.1              &  & 38.0         & ±0.7         & 44.6         & ±0.9        & 61.9           & ±2.8         & 46.0          & ±1.1          \\
\textbf{avg}                  & 41.6 &  & 82.3          &                     &  & 47.5               & 52.5              &  & 48.2         &              & 52.0         &             & 63.8           &              & 53.0          &               \\ \midrule
\textbf{macro avg}            & 38.6 &  & 89.3          &                     &  & 56.6               & 63.2              &  & 61.6         &              & 70.8         &             & 80.7           &              & 72.1          &               \\ \bottomrule
\end{tabular}
\end{small}
\end{center}
\caption{\textbf{Complete results for each task and system.} When applicable, results show fine-tuning average performance and standard deviation over 3 seeds. Full-shot shows the fine-tuning results with the hyperparameters described in Table \ref{tab:appendix-hyperparameters}. Prompting shows the single and multi-verbalizer baseline results. Mining results show single and multi-verbalizer performance without filtering, in addition to multi-verbalizer performance with full-shot and zero-shot filtering. Maj: majority baseline}
\label{tab:complete-results-appendix}
\end{table*}

\begin{table*}[t]
\begin{center}
\begin{small}
\addtolength{\tabcolsep}{-2.5pt}
\begin{tabular}{@{}ll@{}Tl@{}Tl@{}Tl@{}Tl@{}Tl@{}}
\toprule
                    & \multicolumn{2}{l}{\textbf{Amazon}} & \multicolumn{2}{l}{\textbf{IMDB}} & \multicolumn{2}{l}{\textbf{MR}} & \multicolumn{2}{l}{\textbf{SST-2}} & \multicolumn{2}{l}{\textbf{Yelp}} &  \textbf{avg} \\ \midrule
\textbf{Prompting}  &                  &                  &                 &                 &                &                &                  &                 &                 &                 & \\
\textit{good / bad }           & 81.5             &                  & 78.4            &                 & 71.1           &                & 77.3             &                 & 81.9            &                 & 78.1 \\
\textit{great / awful}         & 82.9             &                  & 82.7            &                 & 80.8           &                & 82.6             &                 & 82.6            &                 & 82.3\\
\textit{awesome / terrible}    & 84.5             &                  & 82.0            &                 & 78.3           &                & 82.8             &                 & 84.0            &                 & 82.3\\
\textit{incredible / horrible} & 86.6             &                  & 83.5            &                 & 78.0           &                & 80.0             &                 & 87.2            &                 & 83.1\\
combined                       & 83.5             &                  & 81.8            &                 & 78.3           &                & 81.9             &                 & 83.1            &                 & 81.7\\ \midrule
\textbf{Mining}                &                  &                  &                 &                 &                &                &                  &                 &                 &                 & \\
\textit{good / bad  }          & 71.7             & ±2.2             & 78.2            & ±0.5            & 70.5           & ±1.0           & 77.3             & ±1.4            & 62.1            & ±2.3            & 72.0 \\
\textit{great / awful}         & 88.4             & ±1.7             & 75.7            & ±4.0            & 73.0           & ±2.5           & 79.8             & ±2.1            & 93.3            & ±0.5            & 82.1\\
\textit{awesome / terrible}    & 91.4             & ±0.6             & 81.3            & ±1.4            & 73.7           & ±1.6           & 78.8             & ±1.3            & 94.3            & ±0.7            & 83.9\\
\textit{incredible / horrible} & 90.5             & ±0.2             & 88.1            & ±0.2            & 80.5           & ±1.0           & 83.5             & ±0.6            & 94.0            & ±0.4            & 87.3\\
combined                       & 90.6             & ±1.7             & 83.0            & ±4.1            & 77.7           & ±2.4           & 83.8             & ±2.4            & 92.0            & ±2.2            & 85.4\\ \bottomrule
\end{tabular}
\end{small}
\end{center}
\caption{\textbf{Verbalizer comparison for sentiment tasks.} For mining, we report average performance and standard deviation over 3 seeds without filtering.}
\label{tab:verbalizer-results-appendix}
\end{table*}

\begin{table*}[t]
\begin{center}
\begin{small}
\addtolength{\tabcolsep}{-2.5pt}
\begin{tabular}{@{}ll|l@{}Tl@{}Tl@{}Tl@{}Tl@{}Tl}
\toprule
\textbf{Pattern}                                      & \textbf{Verbalizer}  & \multicolumn{2}{l}{\textbf{Amazon}}            & \multicolumn{2}{l}{\textbf{IMDB}}             & \multicolumn{2}{l}{\textbf{MR}}              & \multicolumn{2}{l}{\textbf{SST-2}}             & \multicolumn{2}{l}{\textbf{Yelp}}             & \textbf{avg}  \\ \midrule
\textbf{Prompting}                                     &             &      &           &      &           &      &           &      &           &      &           &      \\
\verb|{VERBALIZER} stars: {INPUT}|   & \textit{5 / 1}         & 50.4 &           & 50.0 &           & 50.0 &           & 50.9 &           & 53.7 &           & 51.0 \\
\verb|{INPUT} I {VERBALIZER} it.|    & \textit{love / hate }& 77.6 &           & 73.3 &           & 64.6 &           & 69.8 &           & 80.3 &           & 73.1 \\
\verb|{INPUT} It is {VERBALIZER}.|   & \textit{good / bad } & 81.5 &           & 78.4 &           & 71.1 &           & 77.4 &           & 81.9 &           & 78.1 \\ \midrule
\textbf{Mining}                      &             &      &           &      &           &      &           &      &           &      &           &      \\
\verb|{VERBALIZER} star*. {INPUT}|   &\textit{ 5 / 1}       & 68.7 & ±0.4      & 75.4 & ±2.1      & 70.8 & ±1.7      & 78.3 & ±2.4      & 67.5 & ±5.3      & 72.1 \\
\verb|i {VERBALIZER}*. {INPUT}|      & \textit{love / hate} & 88.9 & ±0.3      & 84.0 & ±1.2      & 79.2 & ±0.8      & 84.0 & ±0.8      & 91.1 & ±0.7      & 85.5 \\
\texttt{(is|was) \{VERBALIZER\}*. \{INPUT\}} & \textit{good / bad}  & 71.7 & ±2.2      & 78.2 & ±0.5      & 70.5 & ±1.0      & 77.3 & ±1.4      & 62.1 & ±2.3      & 72.0 \\ \bottomrule
\end{tabular}
\end{small}
\end{center}
\caption{\textbf{Template comparison.} Performance for three different templates on sentiment tasks comparing prompting and mining without filtering. Additionally, we show standard deviations over three seeds for the mining approach. The verbalizer column shows the verbalizer for the positive and the negative class, respectively.}
\label{tab:template-sensitivity-full}
\end{table*}

\begin{table*}[t]
\begin{center}
\begin{small}
\addtolength{\tabcolsep}{-2.5pt}
\begin{tabular}{@{}ll@{}Tl@{}Tl@{}Tl@{}T@{}} \toprule
                              & \multicolumn{4}{l}{\textbf{Default dropout}}                  & \multicolumn{4}{l@{}}{\textbf{High dropout}}                     \\
                              & \multicolumn{2}{l}{full-shot} & \multicolumn{2}{l}{zero-shot} & \multicolumn{2}{l}{full-shot} & \multicolumn{2}{l@{}}{zero-shot} \\ \midrule
\textbf{Sentiment Analysis}   &               &               &                &              &               &               &               &               \\
Amazon                        & 95.8          & ±0.0          & 91.0           & ±0.4         & 94.4          & ±0.2          & 92.0          & ±0.6          \\
IMDB                          & 94.4          & ±0.2          & 80.1           & ±4.0         & 91.6          & ±0.4          & 86.7          & ±1.3          \\
MR                            & 88.0          & ±0.2          & 76.5           & ±1.1         & 86.3          & ±0.7          & 80.5          & ±1.0          \\
SST-2                         & 92.0          & ±0.5          & 80.2           & ±1.6         & 89.5          & ±0.7          & 85.6          & ±1.1          \\
Yelp                          & 96.6          & ±0.1          & 94.4           & ±0.6         & 95.2          & ±0.6          & 92.0          & ±1.5          \\
\textbf{avg}                  & 93.3          &               & 84.4           &              & 91.4          &               & 87.4          &               \\ \midrule
\textbf{Topic Classification} &               &               &                &              &               &               &               &               \\
AGNews                        & 90.8          & ±0.2          & 77.2           & ±1.0         & 89.5          & ±0.2          & 79.2          & ±0.6          \\
DBPedia                       & 98.8          & ±0.0          & 83.9           & ±0.4         & 97.1          & ±0.3          & 80.4          & ±0.3          \\
Yahoo                         & 72.9          & ±0.1          & 56.6           & ±2.4         & 69.8          & ±0.1          & 56.1          & ±2.1          \\
\textbf{avg}                  & 87.5          &               & 72.5           &              & 85.5          &               & 71.9          &               \\ \midrule
\textbf{NLI}                  &               &               &                &              &               &               &               &               \\
MNLI                          & 77.7          & ±1.0          & 52.9           & ±0.8         & 65.5          & ±0.3          & 50.4          & ±0.4          \\
QNLI                          & 77.0          & ±1.6          & 57.8           & ±0.9         & 70.2          & ±1.1          & 53.2          & ±0.6          \\
RTE                           & 72.3          & ±1.1          & 61.4           & ±2.9         & 57.8          & ±1.3          & 62.6          & ±0.9          \\
SNLI                          & 80.2          & ±0.2          & 49.4           & ±1.0         & 61.9          & ±2.8          & 46.0          & ±1.1          \\
\textbf{avg}                  & 76.8          &               & 55.4           & ±0.4         & 63.8          &               & 53.0          &               \\ \midrule
\textbf{macro avg}            & 86.4          &               & 71.8           &              & 80.7          &               & 72.1          &               \\ \bottomrule
\end{tabular}
\end{small}
\end{center}
\caption{\textbf{Performance with high dropout and default dropout.} These results use our proposed mining method + filtering and compares 2 settings of the hidden layer dropout: the default setting of $0.1$ and the high regularization setting of $0.4$, the value that was found most effective during development experiments on AGNews and SST-2.}
\label{tab:results-normal-dropout}
\end{table*}

\begin{table*}[t]
\begin{center}
\begin{small}
\addtolength{\tabcolsep}{-2.5pt}
\begin{tabular}{@{}lccc@{}}
\toprule
 & \multicolumn{1}{l}{\textbf{Averaging}} & \multicolumn{1}{l}{\textbf{Max}} & \multicolumn{1}{l}{\textbf{Sum}} \\ \midrule
 \textbf{Sentiment Analysis} & & & \\
Amazon                               & 83.5                     & 82.7                    & 83.5                    \\
IMDB                                 & 81.8                     & 81.2                    & 82.1                    \\
MR                                   & 78.3                     & 77.4                    & 78.3                    \\
SST-2                                & 81.9                     & 81.3                    & 81.9                    \\
Yelp                                 & 83.1                     & 82.3                    & 83.1                    \\ \midrule
\textbf{Topic Classification}        & \multicolumn{1}{l}{}     & \multicolumn{1}{l}{}    & \multicolumn{1}{l}{}    \\
AGNews                               & 54.6                     & 53.4                    & 54.6                    \\
DBPedia                              & 51.1                     & 49.1                    & 51.4                    \\
Yahoo                                & 34.1                     & 34.1                    & 34.1                    \\ \midrule
\textbf{NLI}                         & \multicolumn{1}{l}{}     & \multicolumn{1}{l}{}    & \multicolumn{1}{l}{}    \\
MNLI                                 & 46.5                     & 46.4                    & 47.0                    \\
QNLI                                 & 58.2                     & 58.4                    & 57.4                    \\
RTE                                  & 61.4                     & 61.7                    & 60.6                    \\
SNLI                                 & 44.1                     & 42.8                    & 43.7                    \\ \bottomrule
\end{tabular}
\end{small}
\end{center}
\caption{Results for different \textbf{probability aggregation functions} for multi-verbalizer prompting.}
\label{tab:results-prompting-agg}
\end{table*}

\begin{table*}[t]
\begin{center}
\begin{small}
\addtolength{\tabcolsep}{-2.5pt}
\begin{tabular}{@{}lcc@{}}
\toprule
                              & \multicolumn{1}{l}{\textbf{Full-shot}} & \multicolumn{1}{l}{\textbf{Prompting}}  \\ \midrule
\textbf{Sentiment Analysis}   & \multicolumn{1}{l}{}                   & \multicolumn{1}{l}{}                    \\
Amazon                        & 68.7                                   & 63.6                                    \\
IMDB                          & 64.5                                   & 63.6                                    \\
MR                            & 66.1                                   & 63.6                                    \\
SST-2                         & 66.7                                   & 63.6                                    \\
Yelp                          & 69.1                                   & 63.6                                    \\ \midrule
\textbf{Topic Classification} & \multicolumn{1}{l}{}                   & \multicolumn{1}{l}{}                    \\
AGNews                        & 52.3                                   & 31.8                                    \\
DBPedia                       & 25.7                                   & 13.4                                    \\
Yahoo                         & 36.4                                   & 22.2                                    \\ \midrule
\textbf{NLI}                  & \multicolumn{1}{l}{}                   & \multicolumn{1}{l}{}                    \\
MNLI                          & 39.2                                   & 42.0                                    \\
QNLI                          & 52.7                                   & 62.2                                    \\
RTE                           & 50.3                                   & 62.2                                    \\
SNLI                          & 40.8                                   & 42.0                                    \\ \bottomrule
\end{tabular}
\end{small}
\end{center}
\caption{\textbf{Label agreement.} Percentage of examples for which the mined label is equal to the label predicted by a full-shot model or by single-verbalizer prompting.}
\label{tab:label-agreement}
\end{table*}

\begin{table*}[t]
\begin{center}
\begin{small}
\begin{tabular}{@{}rlp{124mm}@{}}
\toprule
\# & \textbf{Mined Label} & \textbf{Mined Example}                                                                                                                                                                                                                                             \\ \midrule
1  & Negative    & So I bought this unit, which said it had the same technical features as the other brand, such as number of channels etc, and this one performed amazing!!                                                                                                 \\
2  & Positive    & The founders of Clickfunnels have focused not only on creating a great internet site for you yet also offering you enough expertise and details to act as an informed company person / entrepreneur.                                                      \\
3  & Positive    & Once home, we began priming.                                                                                                                                                                                                                              \\
4  & Negative    & While you can ’ t view the content on the second page without either logging in or signing up since there ’ s no ‘ x ’ button, there ’ s a trick involving 3D Touch that can help.                                                                        \\
5  & Positive    & That i savored them a long way a great deal more compared to My spouse and i believed i would certainly.                                                                                                                                                  \\
6  & Positive    & Do you have an idea of how broad your vocal range was?                                                                                                                                                                                                    \\
7  & Positive    & Also recently the lovely Megan Washington modelled for our latest transeasonal collection we just shot last week.                                                                                                                                         \\
8  & Positive    & What are 3 things that make you happy?                                                                                                                                                                                                                    \\
9  & Positive    & Be devoted to one another in love.                                                                                                                                                                                                                        \\
10 & Negative    & An unexpected situation arose with my father requiring help for three to four months.                                                                                                                                                                     \\
11 & Positive    & Simply AWESOME!                                                                                                                                                                                                                                           \\
12 & Negative    & I don ’ t disagree with that, however the voices have never been as loud or as many as now about this topic of “anti TPS ”, that is progress, that is a movement, that is what we need to inspire EA to finally listen and do something about it.         \\
13 & Positive    & Adverse drug reactions are based on evaluation of data from pre-marketing phase 2-3 studies and updated based on pooled data from 18 placebo-controlled pre- and post-marketing studies, including approximately 5,000 patients treated with varenicline. \\
14 & Negative    & I have the books from last year and have spoke to the college to get this years as they may be changing to another type.                                                                                                                                  \\
15 & Negative    & I was down-to-my-core terrified.                                                                                                                                                                                                                          \\
16 & Negative    & He didn ’ t win, and our support of him became rather limited when we determined he was not winning.                                                                                                                                                      \\
17 & Positive    & It was in a four-star hotel in the Boca Raton Resort Bungalows.                                                                                                                                                                                           \\
18 & Positive    & I wish my preschool was this nice.                                                                                                                                                                                                                        \\
19 & Negative    & People in Wall Street and other financial services firms should have paid more attention to the data.                                                                                                                                                     \\
20 & Positive    & I guess I am quite transparent.                                                                                                                                                                                                                           \\ \bottomrule
\end{tabular}
\end{small}
\end{center}
\caption{Random sample of mined examples for sentiment analysis.}
\label{tab:sentiment-mined-examples-appendix}
\end{table*}

\begin{table*}[t]
\begin{center}
\begin{small}
\begin{tabular}{@{}rlp{62mm}p{62mm}@{}}
\toprule
\# & \textbf{Mined Label}   & \textbf{Mined Premise}                                                                                                                                                                                                                                                                                                                                   & \textbf{Mined Hypothesis}                                                                                                                                                              \\ \midrule
1  & Neutral       & When seniors and their caregivers develop and enjoy deep friendships, it can bolster feelings of well-being, happiness and security.                                                                                                                                                                                                            & it can help to make activities like exercising and healthy eating more enjoyable.                                                                                             \\
2  & Contradiction & Customer service skills training for staff should include basic customer service skills of active listening, how to handle difficult situations, telephone etiquette, and interpersonal communication tips.                                                                                                                                     & most of all, it should focus on how to build a relationship with a student.                                                                                                   \\
3  & Entailment    & The easy way to remember the arrhythmias most commonly associated with SSS — is to think of what one might expect if the SA node became “sick ”.                                                                                                                                                                                                & there is sinus bradycardia and arrhythmia — sinus pauses (which may be longlasting, ultimately leading to sinus arrest) — and SA nodal block.                                 \\
4  & Contradiction & Primarily due to President Obama ’ s historic announcement, more Americans are visiting Cuba, making a bad situation worse.                                                                                                                                                                                                                     & for those of you with the ability to book your Cuban Rent A Car in advance, all the above official websites still offer availability at the writing of this article.          \\
5  & Neutral       & In 1989, Tufts University School of Medicine created the Minority High School Tutorial PLUS Program to provide local minority / disadvantaged students with access to medical student tutors.                                                                                                                                                   & in 1989, Tufts University School of Medicine received a grant from the National Institutes of Health (NIH) to start the Minority High School Research Apprenticeship Program. \\
6  & Contradiction & In fact, regardless of the cost escalations on those other projects, legislators are doing their job by showing such prudence here.                                                                                                                                                                                                             & when all is said and done, the legislature should approve the project aimed at repairing and upgrading seats and improving lighting and drainage at the facility.             \\
7  & Entailment    & Also, this is the root of consciousness, because consciousness, awareness needs an opposite, a counterpart, a border, to awake at.                                                                                                                                                                                                              & consciousness is a form of pain, originally, definitely.                                                                                                                      \\
8  & Contradiction & Does Amaryl cause hair loss?                                                                                                                                                                                                                                                                                                                    & the use of Amaryl does not cause hair loss.                                                                                                                                   \\
9  & Entailment    & “Golay has got no serious issues.                                                                                                                                                                                                                                                                                                               & he is resorting to such statements for cheap popularity, ” remarked Bhim Dahal, the spokesperson of ruling Sikkim Democratic Front (SDF).                                     \\
10 & Neutral       & The point is, it has to go.                                                                                                                                                                                                                                                                                                                     & I'll be removing a lot of the buttons in favor of textual links, and will probably replace them with a single button promoting Firefox.                                       \\
11 & Contradiction & Most of the times precisely originating from a sincere analysis of the weaknesses, the tension field between opposites and the assembling of cross-functional teams, through the clash of diverse approaches and views, the influence of “career changers ” from other fields, and openness towards the new, the unorthodox, the unpredictable. & without diminishing the importance of diversity, togetherness is what produces the most overwhelming feeling of success.                                                      \\
12 & Entailment    & Beef and poultry safety tips are essential to follow.                                                                                                                                                                                                                                                                                           & an effective and healthy way to lose weight is to get regular exercise and harmful toxins, as soon as you have to eat properly.                                               \\
13 & Neutral       & One vehicle that is widely chosen is a motorbike to carry out daily activities.                                                                                                                                                                                                                                                                 & the matic motor is very easy to operate.                                                                                                                                      \\
14 & Entailment    & Storage companies are often located near major travel routes to make it easier for customers to access the facility.                                                                                                                                                                                                                            & you may see signs for local rental companies in your area at the side of the motorway, or major routes near your location.                                                    \\
15 & Entailment    & Connecting with people about a negative experience often equates to a positive outcome.                                                                                                                                                                                                                                                         & I ’ ve decided to list all the ‘ abnormal ’ things I do but wouldn ’ t usually talk about.                                                                                    \\
16 & Entailment    & A smile does go a long way!                                                                                                                                                                                                                                                                                                                     & March 13, coming soon.                                                                                                                                                        \\
17 & Contradiction & Accordingly, it is a cultural taboo to affirm, “I am Love, ” which is our Authentic Self, the Immanent Divine Essence that we all share.                                                                                                                                                                                                        & the Rishis who wrote the Upanishads realized that Brahman and Atman — as the Absolute and Self, respectively — are One, declaring Tat tvam asi ‘ You are That.                \\
18 & Contradiction & In these countries, ceramic proppants are used mainly in wells with higher closure pressures and other challenging environments.                                                                                                                                                                                                                & ceramic proppants are the leading proppant type in the Chinese and Russian markets.                                                                                           \\
19 & Contradiction & Given her recent prognosis and the fact that she DOES drive us mental with her meltdown and seemingly erratic behaviour sometimes, I really need to count to 10 and not lose my rag with her more than I do.                                                                                                                                    & with her reaction to breakfast this morning and subsequent meltdown, I don't think I could have chosen a more difficult resolution...                                         \\
20 & Entailment    & The Best Khao Soi in Chiang Mai, Thailand, is in a Mall!                                                                                                                                                                                                                                                                                        & I said it … the best Khao Soi in Chiang Mai, Thailand, is in Central Airport Plaza Mall.                                                                                      \\ \bottomrule
\end{tabular}
\end{small}
\end{center}
\caption{Random sample of mined examples for NLI.}
\label{tab:nli-mined-examples-appendix}
\end{table*}

\end{document}